\documentclass[runningheads]{llncs}

\usepackage{eccv}

\usepackage{eccvabbrv}

\usepackage{graphicx}
\usepackage{booktabs}
\usepackage[noend]{algpseudocode}
\usepackage{algorithm}
\usepackage{amsmath}
\usepackage{pifont}
\usepackage{wrapfig}
\usepackage{hyperref}
\usepackage[accsupp]{axessibility}  

\usepackage{hyperref}

\usepackage{orcidlink}
\usepackage{tabularx}

\begin{document}

\title{Few-Shot Image Generation by Conditional Relaxing Diffusion Inversion}


\author{Yu Cao\orcidlink{0009-0003-9128-1775} \and
Shaogang Gong\orcidlink{0000-0001-8156-2299}}

\authorrunning{Cao et al.}

\institute{Queen Mary University of London, UK\\
\email{\{yu.cao, s.gong\}@qmul.ac.uk}\\}

\maketitle

\begin{abstract}
In the field of Few-Shot Image Generation (FSIG) using Deep Generative Models (DGMs), accurately estimating the distribution of target domain with minimal samples poses a significant challenge. This requires a method that can both capture the broad diversity and the true characteristics of the target domain distribution. We present \textit{Conditional Relaxing Diffusion Inversion} (CRDI), an innovative `training-free' approach designed to enhance distribution diversity in synthetic image generation. Distinct from conventional methods, CRDI does not rely on fine-tuning based on only a few samples. Instead, it focuses on reconstructing each target image instance and expanding diversity through few-shot learning. The approach initiates by identifying a \textit{Sample-wise Guidance Embedding} (SGE) for the diffusion model, which serves a purpose analogous to the explicit latent codes in certain Generative Adversarial Network (GAN) models. Subsequently, the method involves a scheduler that progressively introduces perturbations to the SGE, thereby augmenting diversity. Comprehensive experiments demonstrates that our method surpasses GAN-based reconstruction techniques and equals state-of-the-art (SOTA) FSIG methods in performance. Additionally, it effectively mitigates overfitting and catastrophic forgetting, common drawbacks of fine-tuning approaches. Code is available at \href{https://YuCao16.github.io/CRDI/}{GitHub}.
\keywords{Few-shot Learning \and Diffusion Model \and Implicit Latent Space}
\end{abstract}

\section{Introduction}
\label{sec:intro}

Deep Generative Model (DGM) has been developed for generating images \cite{higgins2016beta, karras2019style, saharia2022photorealistic}, audio \cite{oord2016wavenet, kong2020diffwave} and point clouds \cite{yang2019pointflow, li2018point}. A notable limitation, however, is their dependency on large-scale datasets and substantial computational resources for optimal performance. 
In many practical applications, only a few samples, sometimes a single sample, are available, \eg photos of rare animal species and some medical images, in which case conventional DGM models are significantly limited \cite{abdollahzadeh2023survey, ojha2021few}. To overcome this problem, Few-Shot Image Generation (FSIG) methods have been proposed \cite{wang2018transferring, mondal2022few, zhao2022few} to generate sufficiently high quality and diverse images with only a few samples as training data, \eg 10 samples. A natural way to achieve this goal is to transform the problem into a few-shot `style' transformation, adapt prior knowledge from generative models built on larger but `similar' source datasets \cite{mondal2022few}. Generative Adversarial Network (GAN) \cite{goodfellow2014generative} is the most widely used method due to its high quality generation. However, if only a few samples are available for learning the underlying distribution of a target domain, standard knowledge transfer approach used in GANs such as fine-tuning suffers from overfitting, mode collapse and catastrophic forgetting \cite{saito2017temporal, radford2015unsupervised, kirkpatrick2017overcoming}.
\begin{figure}[tb]
  \centering
  \includegraphics[width=10cm]{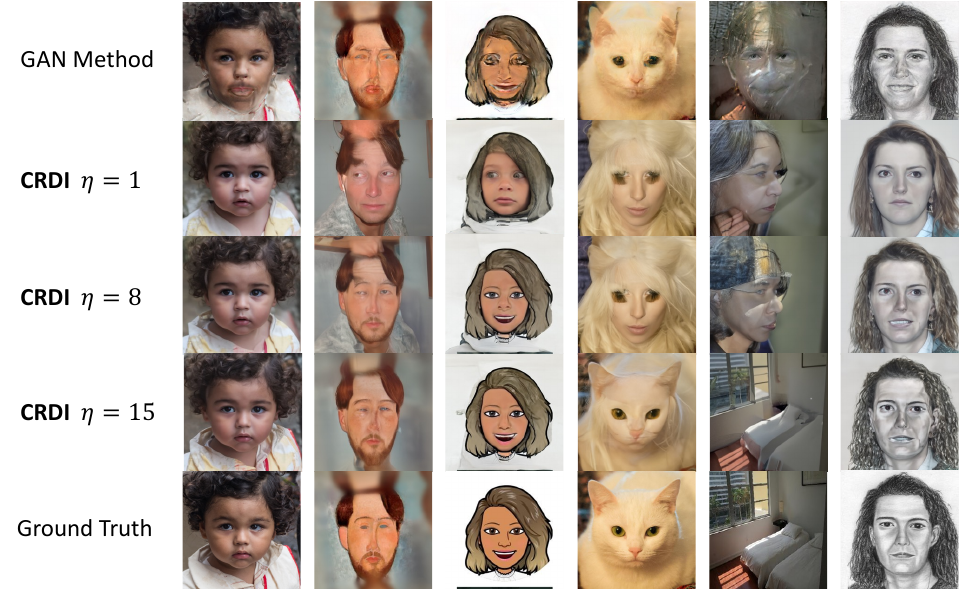}
  \caption{Comparison of reconstruction results using Image2StyleGAN (GAN Method) \cite{abdal2019image2stylegan} and our proposed method (CRDI) with varying $\eta$ values. Both source models are pre-trained on FFHQ \cite{karras2019style}. Our CRDI used the fast sampling method DDIM \cite{song2020denoising} with total 25 inference steps. We further set $\eta\!=\!1, 8, 15$, whilst larger value means a stricter diffusion time-dependent SGE.}
  \label{fig:reconstruction}
\end{figure}

More recently, diffusion models (DMs) \cite{ho2020denoising, song2020score} have demonstrated remarkable success, surpassing GANs in image generation \cite{dhariwal2021diffusion}. In particular, their stochastic processes and probabilistic nature make diffusion models inherently well-suited for tasks such as image generation, text-to-image translation \cite{saharia2022photorealistic, ramesh2021zero, rombach2022high}, and image editing \cite{meng2021sdedit}. It is attractive to consider if DMs can also be developed for FSIG to provide a better solution than the existing methods dominated by GANs. However, directly applying existing adaptation techniques used in GANs including regularization \cite{li2020few, ojha2021few} and modulation \cite{zhao2022few} to the DMs not only fails to solve the problems faced by GANs, but also makes the overfitting and catastrophic forgetting problems even worse due to the significantly larger number of parameters of DMs~\cite{abdollahzadeh2023survey}.
In parallel, it has been shown recently that high-fidelity StyleGAN2 \cite{karras2020analyzing} can represent a latent space for `accommodating' a vast array of out-of-distribution data, as shown in Fig.~\ref{fig:reconstruction} \cite{abdal2019image2stylegan, mondal2022few}. This has triggered a desire to discover and learn a GAN latent space for the target domain therefore enabling image generation by sampling latent codes from the latent space \cite{abdal2019image2stylegan, mondal2022few}, avoiding the pitfalls of overfitting and catastrophic forgetting associated with fine-tuning pre-trained models. However, this approach introduces challenges, such as data leakage, and fails to apply directly to DMs, which lack a deterministic explicit latent code. Current latent space analysis of DMs focus primarily on semantic understanding, relying heavily on extensive data to identify a latent space for controlling the output, through whether learning an embedding space via VAEs \cite{kingma2013auto} or geometric methods, hence incompatible with few-shot settings. \cite{kwon2022diffusion, zhu2023boundary, song2024flow, pandey2022diffusevae}.

To address this problem, we introduce a novel `training-free' approach, \textit{Conditional Relaxing Diffusion Inversion} (CRDI), to maximize distribution diversity in a diffusion generative process, guided by discovering a \textit{Sample-wise Guidance Embedding} (SGE) as a diversity guidance for a target domain. In particular, we discard the traditional concept of \textit{style} transformation and solve the FSIG problem from the perspective of improving diversity. This can be further decomposed into two steps, Reconstruction and Diversity Enhancement. Specifically, \textbf{first}, we discover and estimate an SGE for the diffusion model, which serves as a guidance for the generation path, allowing us to reconstruct a given target sample. Crucially, we allow flexibility in the intermediate noisy states, introducing a conditional relaxing process that enables more robust reconstruction with initial variability. However, we find that the diversity enhancement from this step alone is limited. Inspired by Sadat et al. \cite{sadat2023cads}, we identify that the core diversity issue in diffusion models stems from their tendency to consistently associate identical labels with specific regions in the distribution space through label embeddings. Consequently, what should be stochastic variability is reduced to mere slight perturbations, effectively minimizing randomness to an almost negligible factor. To address this limitation and further boost diversity, we introduce a crucial second step in our approach. \textbf{Second}, by manipulating the SGE through an annealing noise perturbation scheduler, we enhance the diversity of these reconstructions, fulfilling the FSIG objectives without the need for additional training or fine-tuning within the target domain.
This dual-step approach ensures stable estimation of the latent representation for each target sample while enriching the diversity of the target domain for FSIG tasks. To the best of our knowledge, this work is the first to successfully adapt DMs for few-shot domains, bypassing traditional fine-tuning approaches.

Our contributions are as follows:
(1) Introduce and formulate a novel Sample-wise Guidance Embedding (SGE) as a dynamic guidance mechanism for diffusion models, enabling reconstruction within specific domains. We further show both theoretically and experimentally that this SGE possesses comparable functionalities to the explicit latent codes of GANs. 
(2) Propose a novel approach to FSIG by replacing the conventional style transformation with two separate processes utilizing SGE, consisting of a per target instance reconstruction and a few-shot target domain diversity expansion, without any additional training. 
(3) Explore the correlation between the rigidity and diversity of SGE to quantitatively control and provide insight into its effectiveness across different target domains.

\section{Related Works}

\noindent \textbf{Few Shot Image Generation (FSIG)} 
The objective of FSIG is to generate samples that are both high-quality and varied within a novel domain \cite{wang2018transferring}. Conventional approaches typically apply fine-tuning a Generative Model pre-trained on a large dataset of a similar domain \cite{bartunov2018few, wang2018transferring, clouatre2019figr}. However, fine-tuning a full generative network mostly results in overfitting \cite{hu2023phasic}. In practice, fine-tuning only updates a part of a model, \eg BSA \cite{noguchi2019image} and FreezeD \cite{mo2020freeze}. To further improve the effectiveness of fine-tuning given a few shots, EWC~\cite{li2020few}, AdAM~\cite{zhao2022few} and RICK~\cite{zhao2023exploring} exploited some kernel methods by identifying important weights from the source model using Fisher Information \cite{ly2017tutorial} and preserve those knowledge while fine-tuning. In addition, \cite{ojha2021few} and RSSA \cite{xiao2022few} introduce additional loss functions to keep the structure of the generated target domain distribution close to that of the source domain. Hu~\etal~\cite{hu2023phasic}  modified these loss functions in order for an Large Multimodal Model (LMM) such as CLIP \cite{radford2021learning} to be able to apply to a Diffusion Model. In parallel, a representation learning method GenDA \cite{mondal2022few} was introduced for FSIG by constructing a manifold in a latent space.

\noindent\textbf{Foundation Models} 
Because of the high scalability of a diffusion model, many Large Multimodal Models (LMMs) such as DALL-E \cite{ramesh2021zero} and Stable Diffusion \cite{rombach2022high} have gained significant attention for zero-shot (prompt) generalization. Several few-shot adaptation methods based on these foundation models have emerged as a potential new solutions for FSIG, such as DreamBooth \cite{ruiz2023dreambooth}, LoRA \cite{hu2021lora}, and Textual-Inversion \cite{gal2022image}. Although these methods can generate samples from a few shots, they are limited to adapting at the subject level. For optimal performance, the categories of the provided samples must be familiar to the model. Due to computational resource constraints, it is impractical to endlessly expand datasets to cover all target domain categories. Moreover, a fundamental premise of FSIG is that the target and source domains must not overlap \cite{abdollahzadeh2023survey, wang2020minegan, li2020few, zhao2022few}, however, entire target domain may have been exposed during training. Therefore, the research on FSIG remains uniquely challenging.

\noindent\textbf{Generative Model Latent Space}
It has been shown that high-fidelity model such as StyleGAN2~\cite{karras2020analyzing} can capture a latent space for accommodating out-of-distribution image generations~\cite{abdal2019image2stylegan}, see examples in the first row of Fig.~\ref{fig:reconstruction}. To explore this idea, GenDA~\cite{mondal2022few} was proposed to explicitly construct a target data manifold in a GAN latent space in order to generate images by sampling latent codes from this discovered manifold. However, extending this concept to the diffusion model presents greater challenges due to its iterative nature and the absence of a deterministic explicit latent space \cite{ho2020denoising, xiang2023denoising, preechakul2022diffusion}.  The current latent space analysis of diffusion methods is all semantically manipulable using inversion techniques \cite{song2020denoising, wallace2023edict}. These techniques can be summarized in two directions: Some methods~\cite{song2024flow, yang2023diffusion, pandey2022diffusevae} use a variational autoencoder (VAE) \cite{kingma2013auto} to construct implicit latent code and disentangle the desired feature for downstream tasks. Other methods~\cite{zhu2023boundary, kwon2022diffusion} use geometric analysis or a pre-trained LMM such as CLIP to influence the original source domain latent space. These approaches seek to establish an extensive semantic embedding space, necessitating large datasets. 
Our approach is designed to overcome the inherent limitations of constructing a diffusion FSIG model from a small sample size target domain, capable of simultaneously achieving diverse image generations beyond categories closely similar to the source domain, and sufficiently robust tractable model behavior.

\section{Methodology}

\subsection{Preliminaries}

\textbf{Overview} 
Our concept of decomposing FSIG into discrete steps is inspired by the Latent Diffusion Model~\cite{rombach2022high}, showing the advantage of training a separate compression network and a latent code sampler in isolation. This divide-and-conquer principle not only provides a clearer understanding on which components are limiting generation quality and diversity but also simplifies the overall diversification discovery with only a few-shot from the target domain.\\
\textbf{Diffusion Model Background} 
Generating noise from data is a `simple' process, which can be described by the Stochastic Differential Equation (SDE):
\begin{equation}
    d \mathbf{x}_t=\boldsymbol{f}\left(\mathbf{x}_t, t\right) d t+\boldsymbol{g}\left(\mathbf{x}_t, t\right) d \mathbf{w}_t, \quad t \in[0, T]
    \label{eq:forward_sde}
\end{equation}
where $\mathbf{w}$ is the standard Wiener process, $\boldsymbol{f}: \mathbb{R} \rightarrow \mathbb{R}$ and $\boldsymbol{g}: \mathbb{R} \rightarrow \mathbb{R}$ are scalar drift and diffusion coefficients, respectively, with discrete time variable $t \in [0, T]$. Song \etal \cite{song2020score} shows that the SDE in Eq.(\ref{eq:forward_sde}) can be converted to a generative model by first sampling $x_T \sim \mathcal{N}\left(\mathbf{0}, \mathbf{I}\right)$ and then reversing the process through another SDE:
\begin{equation}
    d \mathbf{x}_s = \left[-\boldsymbol{f}\left(\mathbf{x}_s, s\right) + \boldsymbol{g}\left(\mathbf{x}_s, s\right) \boldsymbol{g}\left(\mathbf{x}_s, s\right)^{\top} \nabla_{\mathbf{x}_s} \log p_{s}\left(\mathbf{x}_s\right)\right] dt + \boldsymbol{g}\left(\mathbf{x}_s, s\right) d\mathbf{w}_s
    \label{eq:backward_sde}
\end{equation}
which is the reverse-time Eq.(\ref{eq:forward_sde})~\cite{anderson1982reverse, haussmann1986time, song2020score}, where $x_s := x_{T-t}$ and $\nabla_{\mathbf{x}_s} \log p_{s}\left(\mathbf{x}_s\right)$ is the score function of the marginal distribution over $x_s$. Correspondingly, $\boldsymbol{f}$ and $\boldsymbol{g}$ are chosen to satisfy $x_T \sim \mathcal{N}\left(\mathbf{0}, \mathbf{I}\right)$. Further more, thanks to Tweedie’s formula \cite{stein1981estimation, efron2011tweedie}, score network $\mathbf{s}_{\boldsymbol{\theta}}\left(\mathbf{x}_t, t\right)$ can be proven to be equivalent to a noise network $\mathbf{\epsilon}_{\boldsymbol{\theta}}(\mathbf{x}_t, t)$ introduced in Denoising Diffusion Probabilistic Model (DDPM) \cite{ho2020denoising}, which describes diffusion process from probability viewpoint. The reverse process equation using the noise prediction network is given by:
\begin{equation}
   x_{t-1} = \sqrt{\bar{\alpha}_{t-1}}\left(\frac{x_t-\sqrt{1-\bar{\alpha}_t} \epsilon_\theta(x_t, t)}{\sqrt{\bar{\alpha}_t}}\right)+\sqrt{1-\bar{\alpha}_{t-1}} \epsilon_\theta(x_t, t).
   \label{eq:predict_p}
\end{equation}

\noindent\textbf{Problem Definitions}
To ensure notation consistency and enhance clarity, we formulate the FSIG task as follows: Consider a pre-trained generative model, the underlying distribution of the source data on which the model was trained is represented as $P_S$.  Given a few samples from a target domain $\mathcal{T}$ with underlying distribution $P_T$, the goal is to leverage the pre-trained generative model to synthesize samples $x \sim P_T$, effectively approximating the target distribution. Here, $\mathcal{S}$ and $\mathcal{T}$ represent the source domain and target domain respectively. Unless specified otherwise, the number of given samples of $\mathcal{T}$ is set to 10.

\begin{figure}[tb]
  \centering
  \includegraphics[width=11cm]{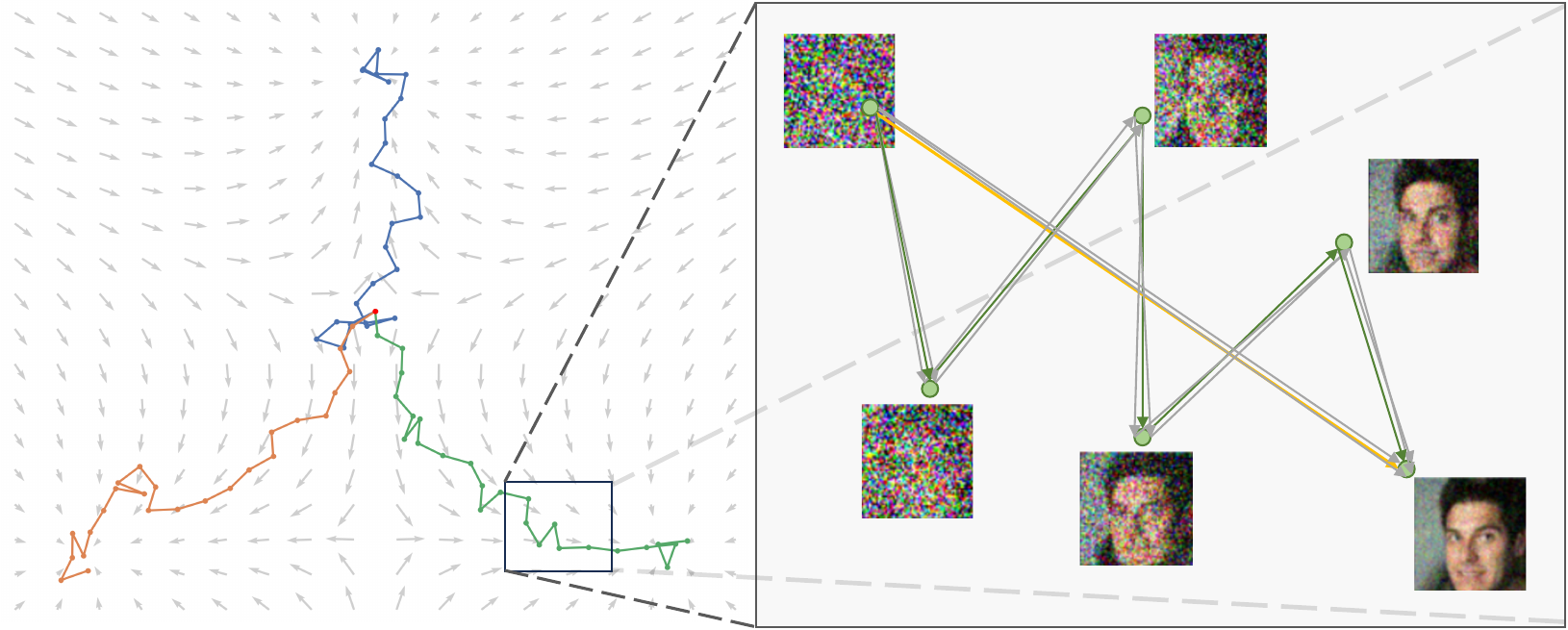}
  \caption{A visualization of three randomly sampled trajectories (blue, orange and green), all originating from the same initial point (red) and generated using Langevin dynamics. The green dot represent the distribution of the intermediate state $x_t$. A time-independent SGE is learned from one direct trajectory (yellow), which can be regarded as a directional path from $x_\alpha$ to $x_\beta$. The SGEs used to guide generation are perturbed by noise (grey) as defined in Sec.~\ref{sec:diversity}. Note that the right corner does not represent $x_0$.}
  \label{fig:trajectory}
\end{figure}

\subsection{Unseen Target Domain Reconstruction}
\label{sec:reconstruct}
Unseen target domain reconstruction can be regarded as a fine-grained conditional generating. A conditional generative model can be derived as $p_t(\mathbf{x}(t) \mid \mathbf{y})$ where $\mathbf{y}$ is the condition.   Per Bayes' theorem, $p_t(\mathbf{x}(t) \mid \mathbf{y}) \propto p_t(\mathbf{x}(t)) p(\mathbf{y} \mid \mathbf{x}(t))$, express this relationship to a score function, a conditional DM is described as:
\begin{equation}
    \nabla_{\mathbf{x}} \log p_t(\mathbf{x}(t) \mid \mathbf{y})=\nabla_{\mathbf{x}} \log p_t(\mathbf{x}(t))+\nabla_{\mathbf{x}} \log p(\mathbf{y} \mid \mathbf{x}(t))
    \label{eq:score_bayes}
\end{equation}
where $\nabla_{\mathbf{x}} \log p_t(\mathbf{x}(t))$ and $\nabla_{\mathbf{x}} \log p(\mathbf{y} \mid \mathbf{x}(t))$ are respectively the scores of a unconditional diffusion model and a time-dependent classifier \cite{song2020score}. Eq.(\ref{eq:score_bayes}) offers a transformation from unconditional to conditional sampling. Rather than using a module with simpler structures trained by few target domain samples to act as the classifier, we claim that the latter form does not necessarily to be a trainable module, instead, it can be a fixed sample-wise embedding, which guides the generating process, we call it a Sample-wise Guidance Embedding (SGE). Fig.~\ref{fig:trajectory} shows a diagram of the proposed SGE. Since the classifier~(Eq.(\ref{eq:score_bayes})) is time-dependent, we further classify the SGE vector into two forms: time-dependent $G_\theta(t)$ and, time-independent $G_\theta$ which is a special case of $G_\theta(t)$.

\begin{algorithm}[tb]
\caption{Proposed Method Pseudo Code ($\eta=1$)}
\begin{algorithmic}[1]
\State \textbf{Input:} Target Domain $\mathcal{T}$, Time Parameter $\alpha$ and $\beta$, Randomly Initialized SGE $G_\theta^i$ for $i \in [1, N]$, a Frozen Noise Network $\epsilon_\theta$ and Learning Rate $\nu$.
\While{not converge}
    \For{$i, x_0$ in \textit{enumerate}($\mathcal{T}$)}
        \State Sample $t$ uniformly from $[\beta, \alpha]$
        \State Given $x_{t-1} \leftarrow$ sample from $\sqrt{\bar{\alpha}_{t-1}} x_0 + \sqrt{1 - \bar{\alpha}_{t-1}} \epsilon, \quad \epsilon \sim \mathcal{N}(0, \mathbf{I})$
        \State $\hat{\epsilon} \leftarrow \epsilon_\theta\left(x_t\right)-\sqrt{1-\bar{\alpha}_t} G_{\theta}^i$
        \State $x_{0}' \leftarrow \frac{x_t - \sqrt{1-\bar{\alpha}_t} \hat{\epsilon}}{\sqrt{\bar{\alpha}_t}}$
       \State $x_{t-1}' \leftarrow \sqrt{\bar{\alpha}_{t-1}}\left(\frac{x_t-\sqrt{1-\bar{\alpha}_t} \hat{\epsilon}}{\sqrt{\bar{\alpha}_t}}\right)+\sqrt{1-\bar{\alpha}_{t-1}} \hat{\epsilon}$
        \State $G_{\theta}^i \leftarrow G_{\theta}^i - \nu \nabla_{G_{\theta}^i} \mathcal{L}$ 
        \State where $\mathcal{L} = \|x_0 - x_0'\|^2 + \| x_{t-1} - x_{t-1}'\|^2$ + $||G_\theta^i - \frac{1}{N}\sum_{j=1}^N G_\theta^j||^2$
    \EndFor
\EndWhile
\State \textbf{return} $G_{\theta}$
\end{algorithmic}
\label{algo:train}
\end{algorithm}

\noindent\textbf{Degree of Rigidity}
For a diffusion model with $T$ inference steps, we introduce a parameter $\eta$, which can take any integer value from 1 to $T$. We define $\eta=T$ as a strict time-dependent SGE and, similarly, $\eta=1$ as a time-independent SGE. We call $\eta$ as a {\em degree of rigidity}. In experiments, we observed that the minimum value of $\eta$ varies across different target domains during reconstruction (Fig.~\ref{fig:reconstruction}). For those images similar to the source domain requires only a small $\eta$ for successful reconstruction. Conversely, for images completely distinct from the source domain, e.g. to reconstruct bedrooms with a model trained on a human face dataset such as FFHQ, even with $\eta = T$ is insufficient for reconstruction. This observation inspired us to consider $\eta$ as a crucial metric for assessing the applicability of the knowledge learned on the source domain to a given target domain. Moreover, from a gradient perspective, as the value of $\eta$ decreases from $T$ to $1$, the role of our SDE shifts from dictating pixel values to influencing the pixel evolution process. More analysis is given in Sec.~\ref{sec:analysis}.

\noindent\textbf{Per Instance Conditional Relaxing Reconstruction}
Pixel-level reconstruction using SGE involves a fixed noisy state $x_t$ obtained by adding noise to $x_0$ according to Eq.(\ref{eq:forward_sde}). However, this condition can be relaxed by allowing the inherent randomness in Eq.(\ref{eq:forward_sde}) to generate different $x_t$ while finding the SGE, thereby enhancing reconstruction diversity.
During the generation process, DMs predict the previous state $x_{t-1}$ from the current state $x_t$ using Eq.(\ref{eq:predict_p}) with a noise prediction network~\cite{ho2020denoising, song2020score}. Additionally, Ho \etal \cite{ho2020denoising} provide a direct estimation of $x_0$ from $x_t$ as shown in Eq.(\ref{eq:predict_s}). In line with our SGE principle, per sample instance model adaptation can be effectively performed using Eq.(\ref{eq:predict_s}). Nonetheless, during the generation phase, the SGE with $\eta > 1$ not only aids in estimating the initial state $x_0$ (Fig. \ref{fig:trajectory}) but also facilitates the generation of the preceding state $x_{t-1}$. Therefore, our loss function is formulated as Eq.(\ref{eq:loss_function}):

\noindent\begin{minipage}{.5\textwidth}
\vspace{-1\baselineskip}
\begin{equation}
    \mathcal{L} = \|x_0 - x_0'\|^2 + \|x_{t-1} - x_{t-1}'\|^2
    \label{eq:loss_function}
\end{equation}
\end{minipage}%
\begin{minipage}{.5\textwidth}
\begin{equation}
    x_{0} = \frac{x_t - \sqrt{1-\bar{\alpha}_t} \epsilon_\theta(x_t, t)}{\sqrt{\bar{\alpha}_t}}
    \label{eq:predict_s}
\end{equation}
\end{minipage}

\noindent where $x_0'$ and $x_{t-1}'$ are derived using Eq.(\ref{eq:predict_s}) and Eq.(\ref{eq:predict_p}), respectively.

\subsection{One-Shot Diversity Enhancement}
\label{sec:diversity}

Diversity is defined as the capacity of a model to generate a variety of outputs for a given condition\cite{sadat2023cads}. As demonstrated by Song \etal \cite{song2020denoising}, abandoning the constraint of a strict Markovian process enables the diffusion process to employ fast sampling by skipping certain steps. This is achieved through a sub-sequence $\tau$ drawn from the sequence $[0,...,T]$.  However, a shorter $\tau$ corresponds to fewer steps in the diffusion process, resulting in reduced diversity of the stochastic processes. Similarly, in our method, varying $\eta$ leads to an equal division of $[0, ..., T]$ into intervals, and for those diffusion processes need greater $\eta$ to reconstruct a target sample, their diversity in the target domain would be negatively affected. To reduce the negative impact of the high degree of rigidity on diversity, we utilize a annealing schedule function $\lambda(t)$ designed by Sadat \etal \cite{sadat2023cads} which corrupts a given condition $y$ based on:

\noindent\begin{minipage}{.5\textwidth}
\begin{equation}
    \hat{\boldsymbol{y}}=\sqrt{\gamma(t)} \boldsymbol{y}+s \sqrt{1-\gamma(t)} \boldsymbol{\epsilon}
    \label{eq:cads_cond}
\end{equation}
\end{minipage}%
\begin{minipage}{.5\textwidth}
\begin{equation}
    \gamma(t)= \begin{cases}1 & t \leq \beta \\ \frac{\beta-t}{\beta-\alpha} & \beta<t<\alpha \\ 0 & t \geq \alpha\end{cases}
    \label{eq:annealing}
\end{equation}
\end{minipage}

\noindent where $\boldsymbol{\epsilon} \sim \mathcal{N}(0, 1)$, $s$ is initial noise scale and $\alpha$ and $\beta$ are the time parameters defining respectively the beginning and the end of the noise scaling interval. In our case, the SGE is the condition $\boldsymbol{y}$ in Eq.(\ref{eq:cads_cond}). 

{\setlength{\intextsep}{10pt}
\begin{figure}[th!]
  \centering
  \includegraphics[width=12cm]{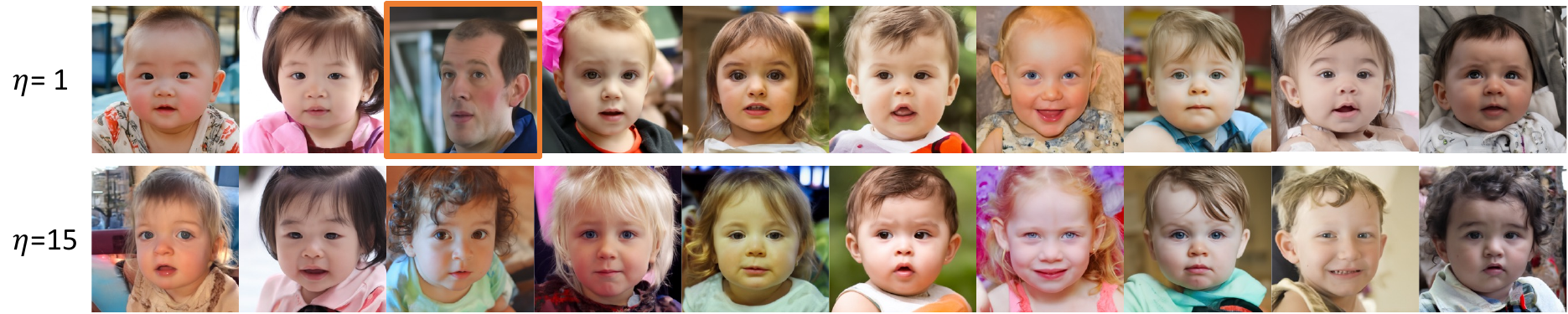}
  \caption{Generated Babies facial images with different $\eta$, slightly
    source domain leakage problem (orange box) when $\eta=1$.}
  \label{fig:eta_babies}
\end{figure}}

However, the SGE in our method is not adaptable based on the current state $x_t$, making the condition scale inapplicable in our case.  Taking this into account, we can rewrite Eq.(\ref{eq:cads_cond}) as: $\hat{\boldsymbol{y}}= \lceil\sqrt{\gamma(t)}\rceil\boldsymbol{y} + s \sqrt{1-\gamma(t)} \boldsymbol{\epsilon}$. Moreover, $\lambda(t) \rightarrow 0$ as $t \rightarrow T$, and $\hat y = s \cdot \epsilon$ is a scaled normal distribution independent of $y$. This allows us to streamline the discovery of the SGE from encompassing all timesteps to focusing on a specific sub-process.  Specifically, the training scheme can be described as: choosing a starting point $\alpha$ and an ending point $\beta \in [0, T]$, for any given sample $x \sim \mathcal{T}$, we first calculate $x_\alpha$ by Eq.(\ref{eq:forward_sde}), then we learn an SGE using pre-trained diffusion model for every $t \in [\beta, ..., \alpha]$. Finally, we add noise perturbation $s \sqrt{1-\gamma(t)} \boldsymbol{\epsilon}$ based on Eq.(\ref{eq:annealing}) to our SGE, as shown by the gray line in Fig.~\ref{fig:trajectory}. However, employing this method on SGE with a low $\eta$ could potentially lead to data leakage problem (Fig.~\ref{fig:eta_babies}) due to the indiscriminate application of noise perturbation. We further discuss this phenomenon in Sec.~\ref{sec:analysis} 

\subsection{Synergy Effect with a Theoretical Analysis}
\label{sec:combine}
In Sec. \ref{sec:reconstruct} and \ref{sec:diversity}, we have described how to construct Sample-wise Guidance Embedding and to increase diversity by Noise Perturbation. Here we explain how we extend this method to the few-shot setting and give a theoretical analysis. From a Score-based Diffusion viewpoint, the goal is to find an explicit solution to solve the reverse-time SDE given by Eq.(\ref{eq:backward_sde}). Specifically, in the probability distribution space $\mathcal{P}$, consider three distributions: the initial distribution $P_I \sim \mathcal{N}(0, I)$; and two distinct final distributions -- a source distribution $P_S$ and a target distribution $P_T$. The transformation from $P_I$ to $P_S$ within the space $\mathcal{P}$ is characterized by Eq.(\ref{eq:backward_sde}) with a pre-trained score network $\mathbf{s}_{\boldsymbol{\theta}}\left(\mathbf{x}_t, t\right)$.

\begin{figure}[th!]
  \centering
  \includegraphics[width=12cm]{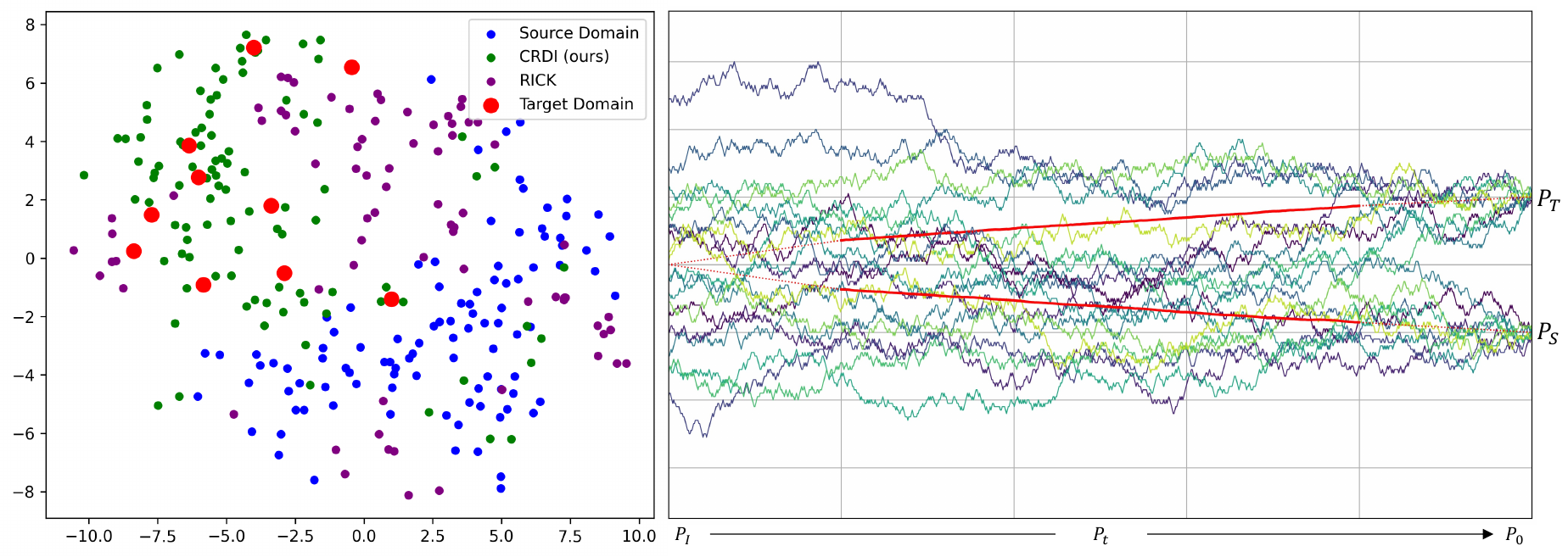}
  \caption{Left: t-SNE results of given samples from Target Domain (Babies) (red), Source Domain (FFHQ) (blue), our generated samples (green), RICK \cite{mondal2022few} generated samples (purple). We show that our generated samples are more align with given target domain samples over RICK. Right: A simulation depicting two SDE transitions from $P_I$ to the $P_S$ and $P_T$. The two solid red lines illustrate the mean trajectories towards the Source and Target Domains, while the red dashed line indicates their extension.}
  \label{fig:sde_tsne}
\end{figure}

In a FSIG task, we want a score network $\mathbf{s}'_{\boldsymbol{\theta}}\left(\mathbf{x}_t, t\right)$ which transforms $P_I$ to the target distribution $P_{T}$. Let $X_t$ represents the state of such a stochastic variable at time $t$. Accordingly, for $\alpha$ and $\beta \in [0, T]$ and $\beta < \alpha$, Eq.(\ref{eq:backward_sde}) forges a dynamic bridge from $X_\alpha$ to $X_\beta$.  Despite the absence of $\mathbf{s}'_{\boldsymbol{\theta}}\left(\mathbf{x}_t, t\right)$, we can utilize Eq.(\ref{eq:forward_sde}) to sample infinite intermediate state $X_t$ for any given sample and hence optimize an SGE as described in Sec.~\ref{sec:reconstruct}. In essence, our SGE can be viewed as an individual trajectory guidance from $P_I$ to $P_T$.
However, it is critical to recognize that learning with a single sample may lead to substantial bias, potentially leading to generated samples falling outside the target domain (implausible diversion). To address this, we employ a mean across all the sample-wise guidance embeddings to serve as a penalty loss. The aggregate mean of the SGE across all given samples serves as an approximation of $\mathbf{s}'_{\boldsymbol{\theta}}\left(\mathbf{x}_t, t\right)$, providing a general guidance for the transition process. This strategy offers a practical solution by transitioning from a sample-wise perspective to a set-wise perspective. Consequently, this `set-wise' guidance embedding ensures a more stable learning process by providing a robust and generalized direction for the transformation. Fig.~\ref{fig:sde_tsne} illustrates an overview of this concept and Algo.~\ref{algo:train} summarizes the overall model learning process.

\section{Experiments}
\label{sec:experiments}

\noindent\textbf{Datasets}
Following previous work \cite{mondal2022few,li2020few, ojha2021few}, we used the Flickr Faces HQ (FFHQ) \cite{karras2019style} as the source domain datasets. We constructed a FSIG diffusion model to adapt to the following common target domains for comparisons to existing FSIG methods: (1) FFHQ-Babies \cite{ojha2021few}, (2) FFHQ-Sunglasses \cite{ojha2021few}, (3) Face Sketches \cite{wang2008face}, (4) Emoji Faces from bitmoji.com API \cite{taigman2016unsupervised}, (5) MetFaces \cite{karras2020training}, (6) portrait paintings from the artistic faces dataset \cite{yaniv2019face}.\\
\noindent\textbf{Metrics and Baseline} 
For the reconstruction task, we calculate the SSIM (Structural similarity index measure) as quantitative metrics. FID (Fréchet inception distance) \cite{heusel2017gans} and Intra-LPIPS (Intra-cluster pairwise Learned Perceptual Image Patch Similarity) \cite{zhang2018unreasonable, ojha2021few} are the most commonly used metrics in FSIG tasks, quantitatively measuring how closely the generated samples match the target domain in terms of quality and diversity, respectively. We further propose a new metric, MC-SSIM (Mode Coverage Structural Similarity Index Measure),  which calculates the average of the top n SSIM scores for each generated image against the given set of target samples, a higher MC-SSIM score indicates superior mode coverage.
We compared our proposed method against 11 FSIG models including TGAN \cite{wang2018transferring}, TGAN+ADA \cite{karras2020training}, BSA \cite{noguchi2019image}, FreezeD \cite{mo2020freeze}, EWC \cite{li2020few}, CDC \cite{ojha2021few}, RSSA \cite{xiao2022few}, DDPM-PA \cite{zhu2022few} AdAM \cite{zhao2022few}, RICK \cite{zhao2023exploring} and GenDA \cite{mondal2022few}. RICK and GenDA are considered the SOTA methods for fine-tuning and representation learning, respectively.\\ 
\noindent\textbf{Implementation Details} 
For the source model, we used Guided Diffusion \cite{dhariwal2021diffusion} and checkpoint from Baranchuk \etal \cite{baranchuk2021label} for FFHQ at 256$\times$256. We further utilized DDIM \cite{song2020denoising} and set the inference step at 25. Model learning was performed on A100 \& H100 GPU with batch size 10. We considered 10 randomly sampled target samples, same as in existing methods for fair comparison, unless otherwise specified. For more details refer to the supplementary material. 

\subsection{Results}
\label{sec:results}

\begin{table}[tb]
\caption{SSIM \cite{wang2004image} Score ($\boldsymbol{\uparrow}$) of Image2StyleGAN \cite{abdal2019image2stylegan} and CRDI (ours) vary $\eta$, quantifying the reconstruction effectiveness.}
\label{tab:reconstruct_ssim}
\centering
\begin{tabular}{cccccccc}
\toprule \textbf{Method} & \textbf{Backbone}  & \textbf{Babies} & \textbf{Amedeo} & \textbf{Bitmoji} & \textbf{Cat} & \textbf{Bedroom} & \textbf{Sketches} \\
\midrule
Image2Style & StyleGAN2 & 0.57 & 0.76 & 0.68 & 0.73 & 0.52 & 0.43  \\
CRDI $\eta=1$ & DDPM & 0.55 & 0.66 & 0.67 & 0.65 & 0.58 & 0.48 \\
CRDI $\eta=8$ & DDPM & 0.68 & 0.75 & 0.81 & 0.77 & 0.67 & 0.63 \\
CRDI $\eta=15$ & DDPM & \textbf{0.74} & \textbf{0.84} & \textbf{0.84} & \textbf{0.83} & \textbf{0.84} & \textbf{0.71} \\
\bottomrule
\end{tabular}
\end{table}

\begin{figure}[tb]
  \centering
  \includegraphics[width=11cm]{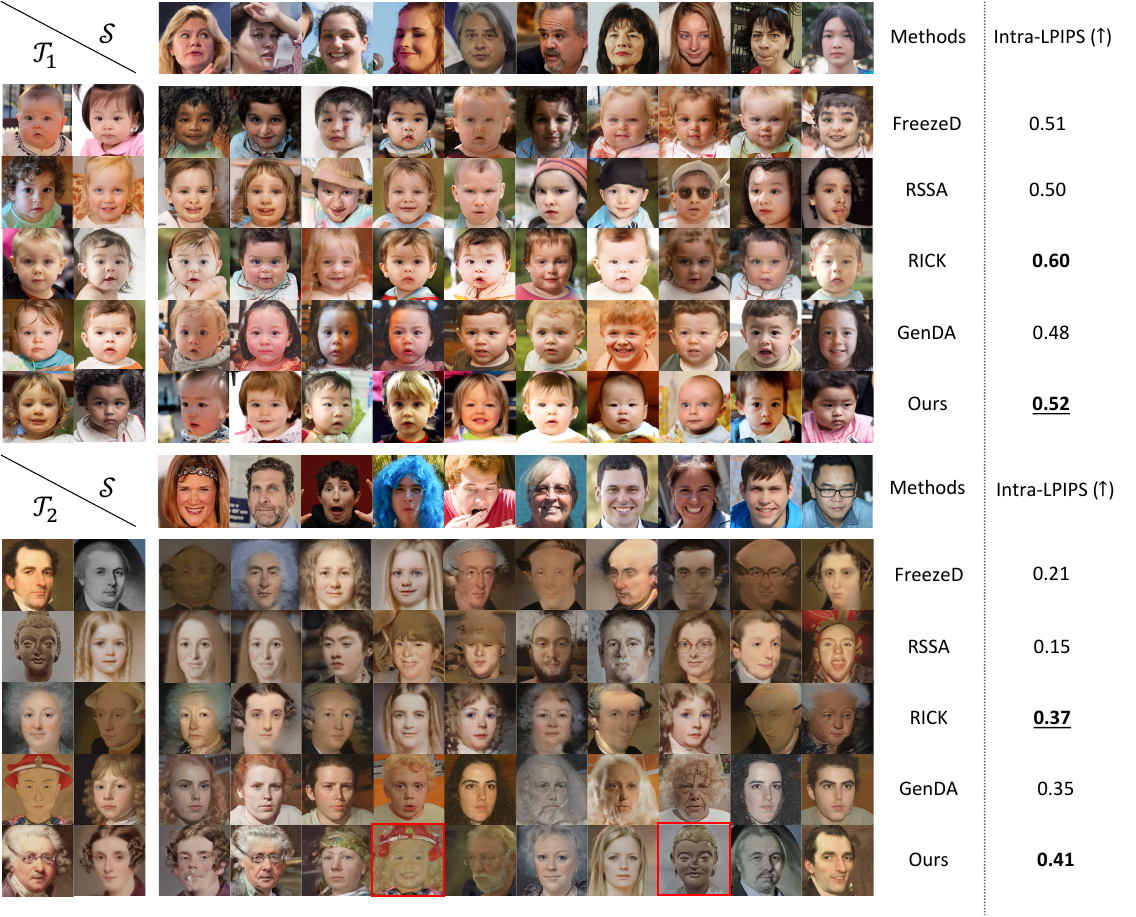}
  \caption{We present the generated samples and Intra-LPIPS ($\boldsymbol{\uparrow}$) for our method alongside four other high performance methods across Babies ($\mathcal{T}_1$) and MetFaces ($\mathcal{T}_2$) with different degrees of similarity to the source domain ($\mathcal{S}$). While not consistently the best in Intra-LPIPS ($\boldsymbol{\uparrow}$), the quality and mode coverage (red box) of our samples is superior, characterized by fewer artifacts and an absence of noticeable overfitting phenomena. Best in \textbf{bold} and the second best in \textbf{\underline{underline with bold}}. For more examples, please refer to Supplementary.}
  \label{fig:result}
\end{figure}

\noindent\textbf{Per Target Instance Reconstruction}
We performed a comparative analysis, both qualitatively and quantitatively, of our reconstruction results against those obtained using Image2StyleGAN \cite{abdal2019image2stylegan}, as shown in Fig. \ref{fig:reconstruction} and Tab. \ref{tab:reconstruct_ssim}. The comparison was carried out on distinct images from six domains with different similarity with the source domain FFHQ (examples can be found in Fig.~\ref{fig:result}). Additionally, in line with the methodology described in Sec. \ref{sec:reconstruct}, we also compared the outcomes with different values of $\eta$. It is evident that our method consistently outperformed GAN-based method (Image2StyleGAN \cite{abdal2019image2stylegan}) across all six domains.  Qualitatively, our technique excelled, especially in the Babies and Bitmoji categories, reconstruction can be achieved even when $\eta=1$ without introducing artifacts. For domains that differ significantly from source domain, such as Bedrooms \cite{yu15lsun} and Amedeo paintings\cite{yaniv2019face}, a larger $\eta$ is required for reconstruction. Surprisingly, although the Sketches \cite{wang2008face} appear to be similar to source domain, they cannot be fully reconstructed, even with strict time-dependent SGE.

\noindent\textbf{FSIG Qualitative Evaluation}
We show examples of the generated images of our method across two target domains (Babies and MetFaces), which vary in their degree of similarity to the source domain, as in the top and bottom of Fig.~\ref{fig:result}, respectively. It can be observed that the fine-tuning approaches (FreezeD, RSSA, RICK) exhibit artifacts and overfitting in the generation within both target domains, while the representation learning approach (GenDA) results in images with limited diversity (low Intra-LPIPS). Our approach surpasses these methods by minimizing visual artifacts through reconstruction and substantially enhancing image diversity with progressively noise perturbation. However, in the Babies category, there remains a diversity gap between our method and RICK. We would like to emphasize that in many cases guaranteeing the generation quality of the resulting image and aligning it with the target domain is more important than diversity, while over-increasing diversity can be dangerous. To verify that our method outperforms existing methods on this point, we employ a t-SNE \cite{van2008visualizing} analysis against RICK on Babies in Fig. \ref{fig:sde_tsne}, it can be seen that our generated distribution shows a higher level of alignment with the samples from the target domain and a reduced alignment with the source domain. This underscores the superior controllability of our method in the generation process. 

\begin{table}[tb]
\caption{Comparing FID ($\boldsymbol{\downarrow}$) Scores and MC-SSIM ($\boldsymbol{\uparrow}$) (for MetFaces only introduced in Sec.~\ref{sec:experiments}) between 10 different models including our proposed method (CRDI). Best in \textbf{bold} and the second best in \textbf{\underline{underline with bold}}}
\label{tab:fid_results}
\centering
\begin{tabular}{lcc|c|cc}
\toprule 
 & & \multicolumn{1}{c}{\textbf{\phantom{a} Babies \phantom{a}}}& \multicolumn{1}{c}{\textbf{Sunglasses}}& \multicolumn{2}{c}{\phantom{aaa}\textbf{MetFaces}\phantom{aaa}}  \\
\cmidrule(lr){3-6}
\textbf{Method} & \textbf{Backbone} & \scriptsize{FID $\downarrow$} & \scriptsize{FID$\downarrow$} & \scriptsize{\phantom{aaa}FID$\downarrow$\phantom{aaa}} & \multicolumn{1}{c}{\scriptsize{MC-SSIM$\uparrow$}} \\
\midrule
TGAN \cite{wang2018transferring} & StyleGAN & $ 104.79 $ & $ 55.61 $ & $ 76.81 $ & $ 0.61 $ \\
TGAN+ADA \cite{karras2020training} & StyleGAN & $ 101.58 $ & $ 53.64 $ & $ 75.82 $ & $ 0.61 $ \\
BSA \cite{noguchi2019image} & StyleGAN &  $ 140.34 $ & $ 76.12 $ & $ - $ & $ 0.69 $ \\
FreezeD \cite{mo2020freeze} & StyleGAN &  $ 110.92 $ & $ 51.29 $ & $ 73.33 $ & $ 0.64 $ \\
EWC \cite{li2020few} & StyleGAN & $ 87.41 $ & $ 59.73 $ & $ 62.67 $ & $ 0.64 $ \\
CDC \cite{ojha2021few} & StyleGAN2 & $ 74.39 $ & $ 42.13 $ & $ 65.45 $ & $ \textbf{\underline{0.70}} $ \\
RSSA \cite{xiao2022few} & StyleGAN2 & $ 75.67 $ & $ 44.35 $ & $ 72.63 $ & $ 0.68 $ \\
DDPM-PA \cite{zhu2022few} & DDPM &  $ 48.92 $ & $ 34.75 $ & $ - $ & $ - $ \\
AdAM \cite{zhao2022few} & StyleGAN2 & $ 48.83 $ & $ 28.03 $ & $ \textbf{\underline{51.34}} $ & 0.65 \\
RICK \cite{zhao2023exploring} & StyleGAN2 & $ \textbf{39.39} $ & $ \textbf{\underline{25.22}} $ & $ \textbf{48.53} $ & $ 0.69 $ \\
GenDA \cite{mondal2022few} & StyleGAN2 & $ 63.31 $ & $ 35.64 $ & $ 104.48 $ & $ 0.35 $ \\
\hline
CRDI (Ours) & DDPM & \textbf{\underline{48.52}} & $ \textbf{24.62} $ & $ 94.86 $ & $ \textbf{0.78} $ \\
\bottomrule
\end{tabular}
\end{table}

\noindent\textbf{FSIG Quantitative Evaluation}
In Tab. \ref{tab:fid_results}, we present complete FID scores, highlighting the performance of our method as superior against that of other representation learning techniques across all three target domains. Our approach outperforms the SOTA fine-tuning method RICK in the Sunglasses category, but faces challenges in the Babies and MetFaces. The notable discrepancy in MetFaces arises from its inherent variety, including sketches, ceramics, and ancient paintings, with an uneven distribution of these sub-domains in the full dataset. 
While other methods may achieve lower FID scores by excelling in dominant sub-domains, they fail to capture the full range of MetFaces variety. Our approach, however, consistently generates samples across all sub-domains within MetFaces (shown in Fig.\ref{fig:result}). To verify this, we utilize MC-SSIM, which assesses the distribution of generated samples across all target sub-domains. The results in Table~\ref{tab:fid_results}, indicate a clear domain coverage advantage of our method over others. Despite not always achieving the lowest FID scores, the broader coverage highlights its effectiveness in handling complex, diverse domains such as MetFaces.

\begin{figure}[tb]
  \centering
  \includegraphics[width=10.7cm]{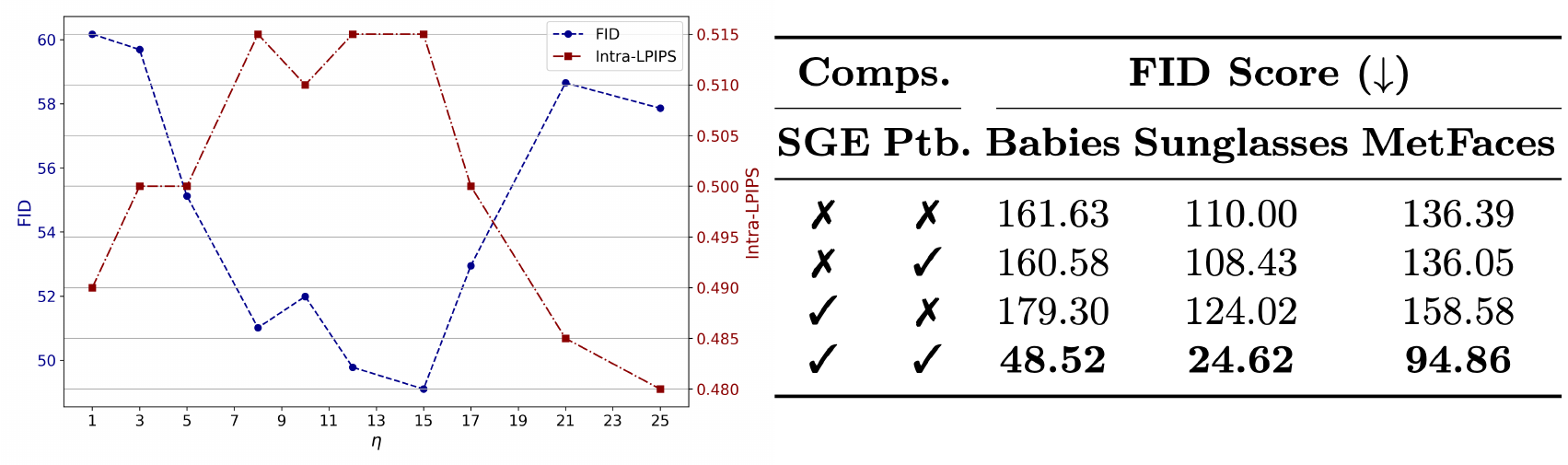}
  \caption{Left: Comparing FID ($\boldsymbol{\downarrow}$) (blue) and Intra-LPIPS ($\boldsymbol{\uparrow}$) (red) across different $\eta$ with target domain Babies. Right: Evaluating the impact of removing each components (Comps.), SGE and noise perturbation (Ptb.) by calculating the FID($\boldsymbol{\downarrow}$) across three target domains.}
  \label{fig:ablation_eta}
\end{figure}

\subsection{Further Analysis and Discussion}
\label{sec:analysis}

\noindent\textbf{Evaluation on Degree of Rigidity} 
In Sec. \ref{sec:reconstruct}, we theoretically analyzed the impact of $\eta$ on the quality and diversity of diffusion generation. In Sec. \ref{sec:results}, we qualitatively and quantitatively analyzed the impact of $\eta$ on reconstruction. Here, we further explore the effects of varying $\eta$ on the generation quality and diversity on Babies, the quantitative results shown in Fig. \ref{fig:ablation_eta} (left). For both FID and Intra-LPIPS, we observe an initial optimization followed by a decline as $\eta$ increases, while the quality of the generated images remains consistent across different $\eta$ values, with minimal artifacts. This pattern can be attributed to a slight data leakage issue when $\eta$ is low, which diminishes as $\eta$ increases (shown in Fig.~\ref{fig:eta_babies}), allowing both FID and Intra-LPIPS to reach their optimal values. However, further increasing $\eta$ imposes a stricter constraint, making the generated images align more with the given samples. This leads to a divergence from the test set distribution, resulting in an increased FID and reduced diversity. This property allows us to adjust between quality and diversity based on actual needs.

Moreover, as shown in Tab.~\ref{tab:reconstruct_ssim} and Fig.\ref{fig:reconstruction}, our good performing category ``Babies'' in terms of FID has the lowest SSIM score during reconstruction. Some fine-grained details (such as the collar) cannot be reconstructed, even when all randomness is removed during generation. We believe that the observed variation can be attributed to the distinct roles that SGE plays in generating samples across various categories, determined by the compatibility of the prior knowledge of the source model with the target domain. For generating images of Babies, SGE serves as a guiding mechanism, as the prior knowledge of the source model aligns with baby images. Conversely, for target domains vastly different from the source domain, it is challenging to apply the prior knowledge learned from the source domain, hence requiring SGE to store more semantic information and thus leaving little room for further diversification in generation. 

\noindent\textbf{Component Effectiveness Evaluation} 
We evaluated the impact of removing each components by calculating the FID scores across three target domains, quantitative results are shown in Fig. \ref{fig:ablation_eta} (Right). It can be observed that the removal of each component has a significant negative effect on model performance measured in FID. Removing SGE, our model is reduced into an unconditional diffusion model, which can only generate samples from source domain. For SGE only, it is reduced to the same setting as for reconstruction only. More visual examples refer to supplementary material.

\noindent\textbf{From Few-Shot to One-Shot}
Till now, our experiments have leveraged 10 images from the target domain for domain adaptation. However, as described in Sec.~\ref{sec:combine}, our method is adaptable with just one image. To demonstrate the effectiveness of our model under this extreme setting, we compared our results with GenDA \cite{mondal2022few} (only compatible method) across Babies and Sunglasses in Tab.~\ref{tab:k_shot}. It is evident that our method outperforms GenDA across all two domains.

\begin{table}[tb]
\caption{Comparisons of model performance from few-shot to one-shot given by the $k$ value in $k$-shot adaptation on generation quality, evaluated by the FID score ($\boldsymbol{\downarrow}$).}
\label{tab:k_shot}
\centering
\begin{tabular}{c c c c c c c}
\toprule
\multicolumn{1}{c}{\textbf{}} & \multicolumn{2}{c}{\textbf{1-shot}} & \multicolumn{2}{c}{\textbf{5-shot}} & \multicolumn{2}{c}{\textbf{10-shot}}\\
\cmidrule(r{4pt}){1-1} \cmidrule(l{4pt}){2-3} \cmidrule(l{4pt}){4-5} \cmidrule(l{4pt}){6-7}
\textbf{Methods} & \textbf{Babies} & \textbf{Sunglasses} & \textbf{Babies} & \textbf{Sunglasses} & \textbf{Babies} & \textbf{Sunglasses}\\
\midrule
GenDA & 105.13 & 83.70 & 65.47 & 45.44 & 62.14 & 35.64 \\ 
Ours & \textbf{100.85} & \textbf{74.60} & \textbf{55.87} & \textbf{31.35} & \textbf{48.52} & \textbf{24.62}  \\ 
\bottomrule
\end{tabular}
\end{table}

\noindent\textbf{Comparison with Foundation Model based Adaptation Methods}

{\setlength{\intextsep}{10pt}
\begin{wraptable}{r}{0.55\textwidth}
\centering
\begin{tabular}{lcccc}
\toprule 
\scriptsize{\textbf{Method}} & \scriptsize{\textbf{Backbone}} & \scriptsize{\textbf{Babies}}& \scriptsize{\textbf{Sunglasses}} & \scriptsize{\textbf{MetFaces}}  \\
\hline
\scriptsize{LoRa} \cite{gal2022image} & \scriptsize{SD-1.5} & \scriptsize{143.78} & \scriptsize{88.38} & \scriptsize{99.65} \\
\scriptsize{DB} \cite{ruiz2023dreambooth} & \scriptsize{SD-2.0} & \scriptsize{172.89} & \scriptsize{160.56} & \scriptsize{187.23} \\
\scriptsize{T-I} \cite{hu2021lora} & \scriptsize{SD-2.0} &  \scriptsize{348.72} & \scriptsize{156.99} & \scriptsize{297.47} \\
\hline
\scriptsize{CRDI} & \scriptsize{DDPM} & \scriptsize{\textbf{48.52}} & \scriptsize{\textbf{24.62}} & \scriptsize{\textbf{94.86}} \\
\bottomrule
\end{tabular}
\caption{Comparing FID ($\boldsymbol{\downarrow}$) of CRDI (ours) vs. DreamBooth (DB), Textual-Inversion (T-I) and LoRa on Babies, Sunglasses and MetFaces.}
\label{tab:fid_foundation}
\end{wraptable}

\noindent 
While foundation models like Stable Diffusion \cite{rombach2022high} can generate diverse images based on prompts, they lack precise control in producing samples that belong to specific domains. Methods based on foundation model such as DreamBooth, LoRA and Textual-Inversion, despite under few-shot settings, are primarily designed for subject-level image editing, resulting in poor performance on FSIG metrics. Moreover, foundation models do not strictly adhere to the FSIG definition, as entire target domain may have been exposed during training. Despite these limitations, our approach significantly outperforms these methods as shown in Tab.~\ref{tab:fid_foundation}, demonstrating superior capability in generating samples that accurately represent the target domain.

}

\section{Conclusion}
In this work, we present a novel framework to tackle the FSIG challenge, showing that limited data can be better utilized through distinct reconstruction and diversity enhancement phases. Our approach achieves SOTA performance using diffusion models, bypassing GANs. This represents a crucial advancement in FSIG technology by offering a measurable balance between the quality and diversity of generated images, and directly assessing the transferability of source models to target domains. Additionally, our method is scalable, compatible with current diffusion models, and optimized for efficiency and lightness.

\noindent\textbf{Limitations \& Future Work}
Our model exhibits reduced effectiveness in reconstructing sketches compared to bedrooms, despite bedrooms being less similar to the RGB facial images (FFHQ).
Looking forward, incorporating CLIP \cite{radford2021learning} as an additional gray-scale image guidance could allow our SGE to focus more on semantic information relevant to the target domain, potentially improving performance across diverse domains without compromising FSIG constraints.

\newpage\noindent\textbf{Acknowledgment.} This work was partially supported by Veritone, Adobe, and utilized Queen Mary’s Apocrita HPC facility from QMUL Research-IT.

%

\bibliographystyle{splncs04}
\bibliography{main}
\end{document}